# Autonomous Last-mile Delivery Vehicles in Complex Traffic Environments

Bai Li[1], Shaoshan Liu[2], Jie Tang[3], Jean-Luc Gaudiot[4], Liangliang Zhang[1*] and Qi Kong[1*]

**Abstract**— E-commerce has evolved with the digital technology revolution over the years. Last-mile logistics service contributes a significant part of the e-commerce experience. In contrast to the traditional last-mile logistics services, smart logistics service with autonomous driving technologies provides a promising solution to reduce the delivery cost and to improve efficiency. However, the traffic conditions in complex traffic environments, such as those in China, are more challenging compared to those in well-developed countries. Many types of objects (such as pedestrians, bicycles, electric bicycles, and motorcycles, etc.) share the road with autonomous vehicles, and their behaviors are not easy to track and predict. This paper introduces a technical solution from JD.com, a leading E-commerce company in China, to the autonomous last-mile delivery in complex traffic environments. Concretely, the methodologies in each module of our autonomous vehicles are presented, together with safety guarantee strategies. Up to this point, JD.com has deployed more than 300 self-driving vehicles for trial operations in tens of provinces of China, with an accumulated 715,819 miles and up to millions of on-road testing hours.

─────────── ◆ ───────────

## 1 Background and Motivations

Logistics service is an essential component of e-commerce for safe and timely delivery of goods to customers. Last-mile delivery here refers to the transportation of goods from a local distribution center to the final recipients. As the final stage of the e-commerce delivery process, autonomous last-mile delivery is extremely challenging as it must handle complex traffic environments.

The main motivations for autonomous last-mile delivery services lies in the intrinsic disadvantages suffered by traditional last-mile deliveries. First, fast growing labor cost may be a serious disadvantage for continued operation and expansion of leading e-commerce companies in China, such as JD.com. Indeed, from our operational data, a contracted delivery clerk with an annual salary of almost $20,000 can deliver 110 e-commerce parcels per day, which renders that each delivery order costs nearly 0.5 USD. This cost is expected to continue increasing as the demographic dividend has reached its end. Secondly, a great portion of working time is wasted as a delivery clerk has to spend time on repeatedly contacting the consumers, waiting for their pick-ups, and traveling on the road, which diverts humans from other more creative work. Fortunately, autonomous driving technologies respond to this exact problem and the benefits of utilizing an autonomous driving vehicle to replace a delivery clerk lie in the following aspects:

1) The delivery is less interrupted by weather conditions or time. Ideally, an unmanned vehicle can respond in a timely manner to the consumers' 24/7 e-commerce orders, which is of particularly importance for late night emergent supplies (e.g. medical) and for people with night work shfits.

2) Similarly, since an unmanned vehicle is designed to operate 24/7, it has more temporal flexibility than the conventional 8-hour working time for deliveries, which provides the customers much more flexibility on the time frame to receive their packages.

3) The costs in terms of labor recruitment, training, and management are greatly reduced.

4) An autonomous vehicle system improves the safety and efficiency of both delivery clerks and other people sharing the public transportation infrastructure. It also effectively prevents the spread of airborn disease, such as SARS and the 2019 Novel Coronavirus, through the interaction between the delivery clerks and customers.

Therefore, deploying autonomous vehicles for the last-mile delivery is a promising approach to overcome the aforementioned disadvantages. In this paper, we will discuss the challenges of autonomous delivery technologies in complex traffic conditions and provide the JD.com's sollution.

## 2. Autonomous delivery technologies in Complex Traffic Conditions

Autonomous driving technologies have been extensively studied in the past few years [1, 2]. However, the traffic conditions in unruly environments are more challenging, primarily because of the large numbers of heterogeneous traffic participants, including pedestrians, bicycles, and automobiles, sharing the roads, and these traffic participants do not necessarily follow traffic rules. These unruly environments are common in various metropolis of China,

───────────
[1] *JDX R&D Center of Automated Driving, JD Inc., Beijing, China & Mountain View, California, U.S.A.*
[2] *PerceptIn, Fremont, California, U.S.A.*
[3] *South China University of Technology, China*
[4] *U.C. Irvine, Irvine, California, U.S.A.*
[*] *Corresponding authors (email: {liangliang.zhang, qi.kong}@jd.com)*

which is distinct from most of the cases in more developed countries, *e.g.,* North America or Europe. Specifically,

1) Since the urban population is large in China, urban residents typically live in apartments rather than individual houses. This naturally means that the population density is high around apartment complexes, which is quite distinct from the situations in more developed countries, which have most of their population in the suburbs. As a result, when traveling in the unstructured environment around apartment environments, an autonomous vehicle would commonly encounter large numbers of complicated interactive objects such as automatic barrier gates in parking lots, pedestrians, bicycle riders, *etc.* (Fig. 1(a)).

2) There are typically multiple types of traffic participants including bicycles, electric bicycles, motorcycles on urban roads, *etc.* Each traffic participant has its own kinematic feature (Fig. 1(b)). To make the situation worse, some residents may use their vehicles in unusual, unsafe or even illegal ways (*e.g.,* the man riding a bicycle while holding an unusually long object in Fig. 1(c), which is actually not a rare scene in areas with dense populations).

3) As depicted in Fig. 1(d), traffic jams often happen in large cities. This is not surprising because motor vehicle ownership keeps growing rapidly.

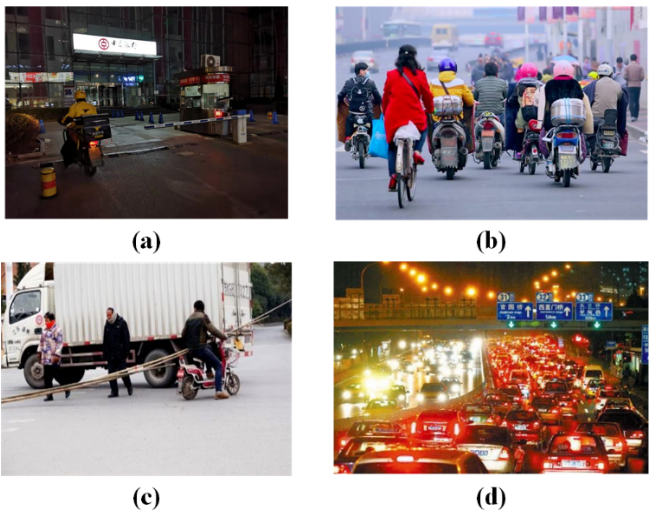

Fig. 1. Typical scenarios that reflect the complexity of unruly driving behaviors: (a) an unstructured parking lot that various traffic tools share the space; (b) urban road with multiple types of two-wheel vehicles; (c) a man who rides carrying with long sticks; and (d) traffic congestion. These factors show that the difference of traffic conditions between more or less developd countries countries (*e.g.,* China).

Compared to other local good transportation systems for commercial applications, such as Nuro, Starship Robot and etc, we target specifically on the traffic conditions in unruly environments (*e.g.,* in some areas of China). The environment is far more complex than those in developed countries, and the delivery vehicle should be able to drive on bycicle lanes, university campus, city lanes, residential areas and etc. Existing autonomous driving solutions suitable for developed countries may not be directly applicable in a chaotic environment. The technologies should be enhanced and adapted to the application scenarios and budget constraints.

Safety should obviously be a primary concern in developing and maintaining autonomous vehicles. Compared with autonomous vehicles that carry passengers, delivery vehicles have unique safety requirements. They should at the same time obey traffic laws and social conventions to make the traffic on the road proceed in a normal, un-blocked, and safe manner. If this is not possible, the delivery autonomous vehicles should not cause an impediment or a danger to the other vehicles, especially other vehicles carrying human passengers. Instead, a delivery vehicle might have to sacrifice itself as an alternative choice, which is a special safety requirement for autonomous delivery vehicles. This design requirement will be discussed in detail in later sections.

## 3 JD.com: An Autonomous Driving Solution

As one of the largest e-commerce companies, JD.com develops autonomous vehicles for last-mile delivery mainly to reduce delivery costs. The cost to deliver each e-commerce parcel would be reduced by approximately 22% if autonomous vehicles were used. This conclusion was made under the assumption that each vehicle can deliver 60 e-commerce parcels per day. If 10% of the entire e-commerce orders in JD.com are delivered by autonomous vehicles, this would result in at least $110 million cost saving annually. More generally speaking, if JD.com can deal with 5% of the e-commerce parcel delivery orders with autonomous driving technologies in the entire market, it would reduce annual costs by a whopping $7.64 billion. The promising benefits and the rational profit mode have been the motivation to carry on the R&D of autonomous vehicles for the last-mile delivery.

### 3.1 Autonomous Driving Architecture

As illustrated in Fig. 2, it takes more than 20 modules working simultaneously and cooperatively to make an autonomous driving system work.

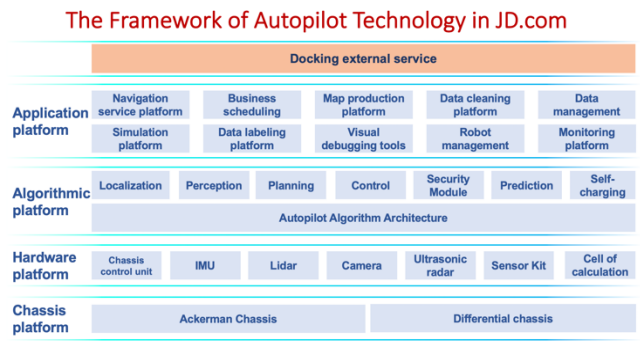

Fig. 2. The architecture of our autonomous driving system.

Broadly speaking, each of the modules shown in Fig. 2 is either an online or an offline module. An online module functions when an autonomous vehicle travels on the road [3], whereas an offline module mainly functions for the purpose of offline feature extraction, training, configuration, simulation, testing and/or evaluation [4].

In a last-mile delivery scheme, a delivery address is assigned to the autonomous vehicle. Thereafter, the current location of the vehicle, as well as the route from the current position to the destination are specified through the navigation service platform. The vehicle then begins to move. During its trip towards the destination, the autonomous vehicle implements localization, local perception, local prediction, local decision, local planning, control, *etc.* in the algorithmic platform to guarantee that the trip is safe, smooth, efficient and predictable.

The hardware platform is a supporting platform for the functioning of the aforementioned elements. The hardware platform consists of the concrete devices, as well as the connections and managements among them. As a general robotic platform, the system is designed for both ackerman steering and differential steering vehicles. The offline modules, on the other hand, prepare for the aforementioned online modules. For example, the map production platform is responsible for the creation of the High-Definition (HD) map; the simulation platform and visual debugging tools help the developers debug. In the remainder of this section, we would like to introduce a few highlighted modules in our autonomous driving technology stack.

### 3.2 Localization and HD Map

Localization is responsible for deriving the current location of the vehicle. In our autonomous driving solution, there are broadly five sources of information which contribute to localization. The first one is the GPS signal, which is used as a startup signal to activate the request of an HD map for later usage.

The remaining four localization sub-modules include localization algorithms based on multi-line Lidar, cameras, chassis-based odometry, and Inertial Measurement Unit (IMU) Information. In using a multi-line Lidar algorithm, the generated point clouds are matched to the pre-stored ones in the requested HD map. The matching is implemented via the generalized Iterative Closest Point (ICP) method [5] online, where the Lidar point clouds are pre-separated into ground and non-ground point clouds.

With multi-line LiDAR, camera, and chassis data at hand, the derivative of the data indicates the movement, which is odometry. Through synthesizing the independent sources of odometry information, a reliable odometry result is derived, otherwise known as as odometry fusion. The fused result, when synthesized with IMU results, is utilized to calibrate the ICP method in a Kalman filter framework [6]. In turn, the ICP method could be further used to calibrate the fusion result so that they could be both closer to the ground truth. Such an intricate architecture is illustrated in Fig. 3.

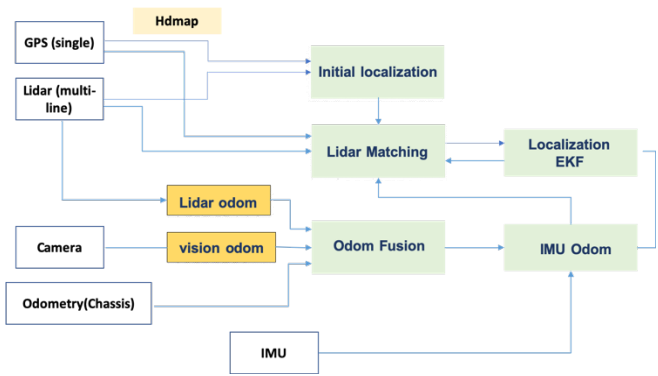

Fig. 3. The architecture of localization.

Our HD map contains the detailed properties of the road, which is represented by 16 layers to provide both static and dynamic information of the environment. They are composed of multiple sub-maps including geometric map, semantic map and real-time map. Some of the elements shared by both autonomous passenger vehicles and delivery vehicles include for example, the location, width, type, curvature, and boundaries of each lane and road, the intersections with their associated semantic features like crosswalks, traffic signals, speed bumps, *etc*. Meanwhile, road elements specifically for delivery vehicles in China are included. For example, detailed representation of pillars (commonly placed on the entrance of bicycle lanes to prohibit large vehicles' access), clear areas (restricted areas where the vehicle cannot stop, could be temporary), gates (gates to access distribution stations or to enter residential communities), safety islands (areas inside a large intersection where pedestrians/cyclists can wait for the next green light) are important in our map. Beside lanes and roads that commonly exist in other HD maps, we added "lane group" elements as an intermediate level between lane and road for better lane associations at large intersections.

Delivery vehicles require cost effective construction of HD maps with the same sensors used on the vehicle, and at all districts in the urban areas, where the GPS signals are usually weak. They also require more frequent updates to HD maps since the roads in urban areas in China change more often than in other places. This requires the establishment of a technical team that focuses on building and maintaining the in-house HD maps based on modern sensor fusion and SLAM technologies that do not rely on a differential GPS or expensive IMUs. Machine learning techniques developed by the perception team are incorporated to assist detection of static vehicles and traffic lights to improve the efficiency of map construction.

### 3.3 Perception

The perception module is responsible for recognizing and tracking the dynamics of the obstacles in the environment. The perception module plays a key role in making the whole autonomous system capable of running through the complex traffic environments. The hardware setup, as illustrated in Fig. 4, is a common multi-source solution. Our vehicle is equipped with one 1-beam LiDAR, one 16-beam LiDAR, four mono cameras to detect the surrounding objects. A high-resolution HDR camera is used for traffic light detection. Ultrasonic receptors are used to prevent immediate collisions with other objects or road elements.

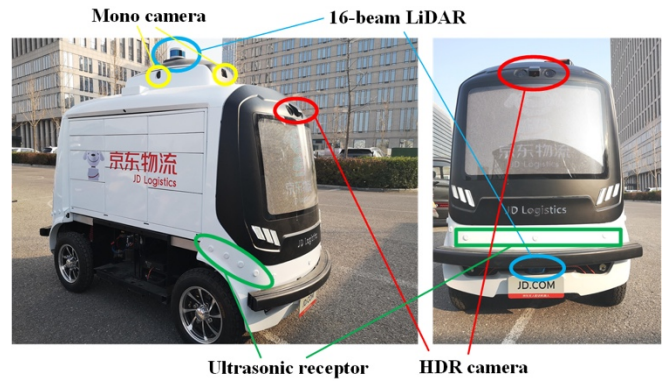

Fig. 4. Sensor hardware setup.

As illustrated in Fig. 5, in object detection, three methods are independently applied, and their detection results are fused. Concretely, the first two detectors process the point cloud data using machine-learning and geometry-based methods, respectively. The third detector processes the visual data using the machine-learning-based methods.

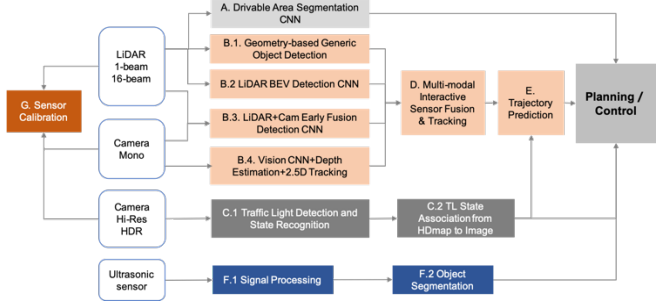

Fig. 5. The architecture of perception.

From our practical experience, we observed that the first method is good at generic traffic-related object classification, but the performance is poor at dealing with pedestrians, which is an inherent drawback of the learning-based methods, because the features of pedestrians are not easy to detect compared to traffic signs. Conversely, the second method provides stable results in recognizing objects with typical shapes (*e.g.*, the pedestrians). The third method is particularly suitable for recognizing partially observed objects because it focuses on color, shading, and texture as identifying features. The three detectors complement each other to provide comprehensive and reliable object recognition results.

As discussed in Section 1, many types of on-road elements exist on the roads and affect autonomous vehicles. Taking the scenario in Fig. 1(c) as an example, the first time our autonomous vehicle encounters such illegal traffic, since the algorithm has not encountered so much data as to correctly realize the relationship between the electric bicycle and the long stick, the ego vehicle (the autonomous delivery vehicle) could be stuck on the road and not know how to handle this scenario. When observing such a case, a remote monitoring human may take over the vehicle as a precaution. Later, this object is placed into a newly established category, and some existing data are marked manually for learning this category in both detectors 1 and 3. We believe that supervised learning-based enumeration is the only feasible way towards achieving reliable recognition performance. In addition, a deduction approach is developed to systematically estimate the status of the interested but unobservable surrounding traffic, such as the status of the traffic lights and that of the vehicle in front of the vehicle right ahead.

### 3.4 Prediction, Decision, and Planning

The prediction, decision and planning module is critical to enabling our autonomous vehicles to safely navigate through complex traffic conditions. Prediction refers to approximating the future trajectories of the tracked moving obstacles. Prediction is done in two cooperative layers. Concretely in the first layer, the behaviors of the well tracked vehicles are predicted with the utilization of the routing information and lane information in the HD map. Herein, we build assumptions on how well the tracked objects will abide by the traffic rules based on the historic road data on this road. The first layer is commonly suitable to handle the regulated vehicles. The second layer targets the prediction of abnormal object behavior, which is achieved through machine learning based methods and deduction methods. The two layers work together to provide the trajectories of the moving objects of interest in the sighting scale of the vehicle. The predicted trajectory is also attached with a variance to measure the prediction confidence.

Decision & planning play a critical role in directly guiding the local driving maneuvers of the vehicle. In dealing with the complex scenarios in a chaotic environment, the decision module should find a rough but reliable homotopic trajectory, while the planning module should run fast. In the decision module, a sample-and-search based method is adopted. In more details, while searching in a graph consisting of sampled nodes, a dynamic programming algorithm is used, and the cost function considers the uncertainties (quantified by confidence degrees mentioned above) in the perception and prediction modules. Through this, a robust and safe trajectory decision is made.

The coarse trajectory derived from the decision module naturally decides whether to yield, or to bypass each of the obstacles. With this coarse trajectory at hand, we can significantly reduce the solution space to a small neighborhood around that coarse trajectory to avoid wasted time in global planning. More specifically, the trajectory planning scheme is split into multiple phases, including decision generation, path planning and velocity planning. Through this decomposition, the original difficulties are largely eased. In the decision generation phase, the interaction between the vehicle and the surrounding objects are determined in a "path" in a path-time coordinate system. Herein, the feasible bounds are inherently forming a tunnel which is determined by the coarse trajectory in the decision phase.

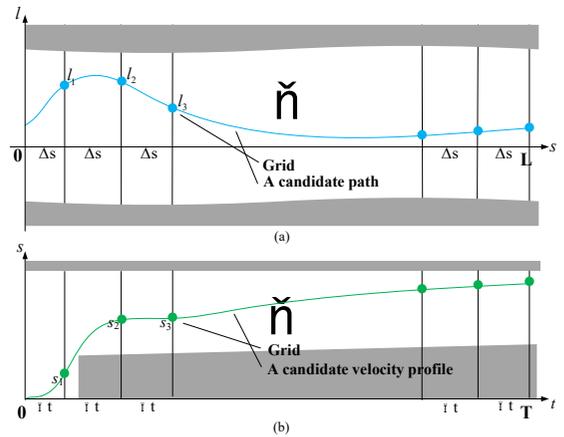

Fig. 6. Schematics on the quadratic programming model formulation: (a) path planning in an *s-l* coordinate system; and (b) velocity planning in an *s-t* coordinate system.

We will discuss this problem in a general format with an *x-y* coordinate system. It can be simply converted to path optimization using a frenet-frame *l-s* (Fig.6(a)), where s is the longitudinal direction along the road center line and l is the lateral shift from the center line, or speed optimization using a path-time frame *s-t* (Fig.6(b)). In more detail, assuming the problem horizon along the *x* axis is fixed as X, we define as many as N straight lines which intersect the *x* axis at (N + 1) equidistant points from 0 to X. With this observation, the path and speed planning schemes can each be converted so as to find the positions of N grids along the lines, so that connecting the grids in a sequence would form a candidate *s-l* path profile or an *s-t* velocity profile. Let us define the position of each grid as $y_i$ ($i=1,…,N$) and the vehicle should obviously not collide with the barriers of the tunnel. The configuration space is therefore formed under basic collision-avoidance constraints. In path optimization, they are formulated by covering the rectangular vehicle as a pair of circles to simplify the calculation, and then require each circle to be collision-free from the two barriers of the tunnel, while for speed optimization, the ego vehicle should not collide with any of the *s-t* regions in the path-time graph [8, 9]. Besides the aforementioned constraints, we expect the trajectory to be smooth, and the optimized result to be as close to the coarse trajectory as it possible. There are other optimization objects and constraints that are related to decisions that the results need to consider. This expectation is reflected in the following minimization objective:

$$J = \sum_k \sum_i \sum_{j=0,1,2,3} w_{k,i,j} \cdot \left(y_i^{(j)} - ref_{k,i,j}\right)^2 \quad (1)$$

In (1), $k \in \{1, ..., N_{ref}\}$ refers to the index of reference profiles. Assumes, and suppose there are is a total of $N_{ref}$ reference profiles which are simultaneously affecting the optimization performance. $i \in \{1, ..., N\}$ denotes the index of grids. $w_{k,i,j} \geq 0$ ($j = 0, ..., 3$) denotes the weighting parameter to encourage the attraction towards the *k*th reference profile at the *i*th grid w.r.t. the *j*th order of derivative of $y_i$ and $ref_{k,i,j}$ ($j = 0, ..., 3$) stands for the corresponding reference profile. As to why we have multiple $ref_{k,i,j}$ at each grid, let us take the velocity planning as an example, at the *N*th grid, n. No one knows *a priori* how to specify the s' at the terminal moment (i.e., $y'_N(N)$). Thus, multiple possibilities are integrated in the form of a weighted sum.

Now with the optimization objective and the constraints at hand, a Quadratic programming (QP) problems are formulated, wherein the cost function is quadratic and the constraints are all linear. We need to minze (3) subject to the dynamic constrains and magnitude constraints at each equidistant point, and the collision constraints from the environment, including static and/or dynamic obstacles. The QP problem can be numerically solved via a local optimizer. The 99[th] percentile time consumption to solve the aforementioned QP problem is within 10 ms, thus the online planning is sufficient to react to the sudden appearance of events and/or of merged objects in complex road scenarios.

## 4 SAFETY AND SECURITY STRATEGIES

As mentioned in previous sections, safety is of paramount importance in designing autonomous delivery vehicles. Safety guarantees are established in multiple layers, including simulation-level, vehicle-end, and remote monitoring. In this section, we introduce a few highlighted safety strategies in our autonomous driving technology stack.

### 4.1 Simulation-level Verification

All submitted codes must be exercised through a large number of benchmark tests. In each test, the input raw data are loaded to "decorate" a virtual real-world test, and the performance of the autonomous vehicle is measured by the virtual output actions in the simulator with a number of carefully defined criteria. In creating the simulation test cases, we utilize the recorded real-world data to build tens of thousands of virtual scenarios. The codes are tested in those virtual but realistic scenarios for performance evaluation. The codes that can pass the simulation tests are tested further in the vehicle-end system, and then the newly emerged issues are recorded and manually marked as virtual scenarios for future code examination. Through this closed loop approach, the safety of the development is rapidly improved.

### 4.2 Vehicle-end Monitoring

We have implemented a vehicle-end low-level guardian module which monitors the occurrence or near-occurrence of an emergency. The guardian module primarily monitors the health of the control system and deals with internal and external exceptions from within the system and from outside. To address the sudden failures that may happen to the hardware, redundant units are equipped to the autonomous vehicles, in association with a failure detection monitor. If even the redundant units fail as well (for example, due to low temperatures that makes the visual sensors fail), the guardian module takes over as per the pre-defined failure-recovery rules. Regarding the troubles from the external environment, when obstacles are overly close to the vehicle, or approaching the vehicle at high speeds, the vehicle would take actions to reduce the collision risks.

### 4.3 Remote Monitoring

To guarantee the operational safety of our delivery vehicles, we have also developed a remote monitoring platform. The driving behavior of the vehicle are monitored in real time. The engineer who remotely monitors the vehicle could take over the control to assist the vehicle to get out of an abnormal situation. If the remotely monitoring engineer is absent, the remote platform would generate a warning signal to inform the police regarding the situation.

## 5 PRODUCTION DEPLOYMENTS

On the way towards large-scale production deployment, we hold a progressive strategy to divide the entire technical difficulties into four stages. The first stage is about autonomously driving at low speeds with manual surveillance. Herein, surveillance stands for the commands from a remote operator for an extra level of safety and asistance. Although direct control by a remote operator is dangerous and impractical due to the network delay and insufficient information provided to the remote site, the human operator can effectively provide useful guidance to the autonomous vehicle. The second stage is about autonomously driving at low speeds without manual surveillance. The third and fourth stages are autonomously driving in relatively high speeds with/without manual surveillance, respectively. Significant efforts in software, hardware, as well as testing are required for the evolution from a lower level to an upper level. On the way to raise the vehicle's driving speed, progress is made adaptively and based on rigorous offline/online testing results.

In addition to the aforementioned progressive technical roadmap, the journal to achieve profitability is also designed to be progressive. When the technologies were far from being appropriate for the on-road trial operations, we focused on developing the low-level chassis, which can be commonly used in many robotic applications. This idea is beneficial in two reasons: (i) The technologies for indoor low-level autonomous moving can be seamlessly applied to warehousing logistics, which will improve the autonomous ability of the full logistic chain; (ii) commercialization and profitability of robotic technogies will arrive before the autonomous driving technologies are ready for full deployment; and (iii) even when the autonomous driving technologies are not mature, the developers should be aware how to develop products, so as to avoid keeping their minds distant from the terminal target, i.e., production.

Regarding the business logic, efforts are also progressively made so that the efficiency of the autonomous last-mile delivery can be maximized. We have been making the scheduling qualities in the e-commerce platform, the warehouse, and the distribution centers take more care about the time efficiency of the deliveries. Improvements in the aforementioned three aspects are made cooperatively and simultaneously to render a good delivery service. Up to this point, we have deployed more than 300 self-driving vehicles for trial operations in several provinces of China, with an accumulated 715,819 miles.

## 6 Lessons Learned

In deploying this autonomous vehicle system, a few lessons have been learned, the first being that the algorithms should be explainable, which make their performance easy to estimate, predict, and thus acceptable for other users sharing the road. Having said this, we find that deep-learning based end-to-end solutions are not pratical at this stage, but Machine-Learning based methods are extensively used in each sub-module with clearly defined boundaries. Secondly, the routes of last-mile delivery vehicles are mostly fixed, so that we heavily rely on HD maps to record fine details along the route. The third lesson is that, during the real-world on-road tests, the pursuit for higher taken-over mile index is misleading because it may lead the developers to hide the risks or problems rather than to find and conquer them. In our viewpoint, accurately recognizing a risk and then request a manual taken-over is highly meaningful, which is a critical part of the entire safety guarantee system. The fourth lesson is that it makes sense to clearly separate the jobs that are suitable for humans and for an automated machine. After a long time of operations in trial, we have learned that it is feasible to allocate the vehicles to handle the complicated but scenarios that repeatedly appear and the human monitors can take over the vehicles when required. In addition, autonomous driving does not mean that human beings become useless, instead they can do innovative work highly related to maintaining an autonomous delivery system with autonomous vehicles.

## Acknowledgments

The authors would like to appreciate the useful discussions and support from Xinyu Xu, Hao Li, Jinfeng Zhang, Shiliang Jin, Wei Dai, Weicheng Zhu, Yahui Cai, Zhuang Li and Haixin Wang. This work was supported in part by the National Key R&D Program of China under Grant 2018YFB1600804.